\documentclass[10pt,twocolumn,letterpaper]{article}

\usepackage{cvpr}
\usepackage{times}
\usepackage[pdftex]{graphicx}
\usepackage{amsmath}
\usepackage{amssymb}
\usepackage{multirow}

\newcommand\blfootnote[1]{%
  \begingroup
  \renewcommand\thefootnote{}\footnote{#1}%
  \addtocounter{footnote}{-1}%
  \endgroup
}

\def\etal{\emph{et al.~}}
\def\ie{\emph{i.e.}}

\cvprfinalcopy 
\ifcvprfinal\fi
\begin{document}
\title{Towards Visually Explaining Variational Autoencoders}

\author{Wenqian Liu$^{1*}$, Runze Li$^{2*}$, Meng Zheng$^{3}$, Srikrishna Karanam$^{4}$, Ziyan Wu$^{4}$, \\ Bir Bhanu$^{2}$, Richard J. Radke$^{3}$, and Octavia Camps$^{1}$\\
$^{1}$Northeastern University, Boston MA   $^{2}$University of California Riverside, Riverside CA\\
$^{3}$Rensselaer Polytechnic Institute, Troy NY   $^{4}$United Imaging Intelligence, Cambridge MA\\
{\tt \scalebox{.7}{liu.wenqi@husky.neu.edu,rli047@ucr.edu,zhengm3@rpi.edu,\{first.last\}@united-imaging.com}}\\
{\tt \scalebox{.7}{bhanu@cris.ucr.edu,rjradke@ecse.rpi.edu,camps@ece.neu.edu}}
}

\maketitle
\thispagestyle{empty}
\begin{abstract}
Recent advances in convolutional neural network (CNN) model interpretability have led to impressive progress in visualizing and understanding model predictions. In particular, gradient-based visual attention methods have driven much recent effort in using visual attention maps as a means for visual explanations. A key problem, however, is these methods are designed for classification and categorization tasks, and their extension to explaining generative models, \eg, variational autoencoders (VAE) is not trivial. In this work, we take a step towards bridging this crucial gap, proposing the first technique to visually explain VAEs by means of gradient-based attention. We present methods to generate visual attention from the learned latent space, and also demonstrate such attention explanations serve more than just explaining VAE predictions. We show how these attention maps can be used to localize anomalies in images, demonstrating state-of-the-art performance on the MVTec-AD dataset. We also show how they can be infused into model training, helping bootstrap the VAE into learning improved latent space disentanglement, demonstrated on the Dsprites dataset.
\end{abstract}

\section{Introduction}
\blfootnote{$^{*}$Wenqian Liu and Runze Li contributed equally to this work.}
Dramatic progress in computer vision, driven by deep learning \cite{krizhevsky2012imagenet,he2016deep,huang2017densely}, has led to widespread adoption of the associated algorithms in real-world tasks, including healthcare, robotics, and autonomous driving \cite{Jin2019AccurateEG, Zuo_2019_CVPR, Li_2019_CVPR} among others. Applications in many such safety-critical and consumer-focusing areas demand a clear understanding of the reasoning behind an algorithm's predictions, in addition certainly to robustness and performance guarantees. Consequently, there has been substantial recent interest in devising ways to understand and explain the underlying \textsl{why} driving the output \textsl{what}.

\begin{figure}[t!]
\centering
\includegraphics[width=\linewidth]{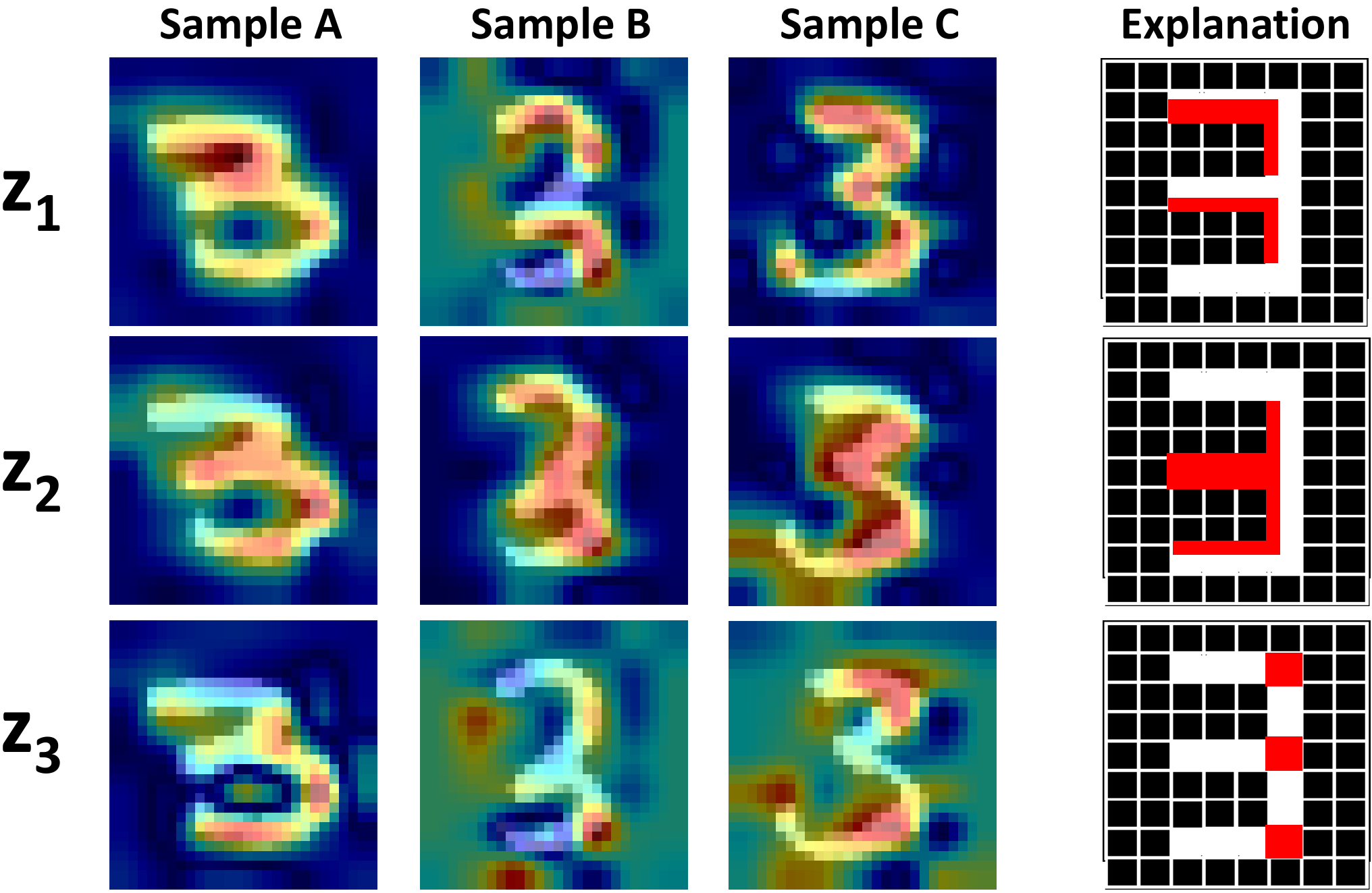}
\caption{We propose to visually explain variational autoencoders. Each element in the latent vector (here $z_1-z_3$) can be explained separately with our attention maps, visualizing consistent explanations across different samples. }
\label{fig:teaser}
\end{figure}

Following the work of Zeiler and Fergus \cite{zeiler2014visualizing}, much recent effort has been expended in developing ways to visualize feature activations in convolutional neural networks (CNNs). One line of work that has seen increasing adoption involves network attention \cite{CAM_CVPR16,Selvaraju_2017_ICCV}, typically visualized by means of attention maps that highlight feature regions considered (by the trained model) to be important for satisfying the training criterion. Given a trained CNN model, these techniques are able to generate  attention maps that  visualize where a certain object, \eg, a \textsl{cat}, is in the image, helping explain why this image is classified as belonging to the \textsl{cat} category. Some extensions \cite{li2019guided, Wang_2019_ICCV} provide ways to use the generated attention maps as part of trainable constraints that are enforced during model training, showing improved model generalizability as well as visual explainability. While Zheng \etal \cite{zheng2019re} used a classification module to show how one can generate a pair of such attention maps to explain why two images of people are similar/dissimilar, all these techniques, by design, need to perform classification to guide model explainability, limiting their use to object categorization problems.

Starting from such classification model explainability, one would naturally like to explain a wider variety of neural network models and architectures. For instance, there has been an explosion in the use of generative models following the work of Kingma and Welling \cite{Kingma2013AutoEncodingVB} and Goodfellow \etal \cite{goodfellow2014generative}, and subsequent successful applications in a variety of tasks \cite{Isola_2017_CVPR, NIPS2017_6644, wang2018vid2vid, Xian_2019_CVPR}. While progress in algorithmic generative modeling has been swift \cite{Wu_2019_CVPR, Kaneko_2019_CVPR, Mehrasa_2019_CVPR}, explaining such generative algorithms is still a relatively unexplored field of study. There are certainly some ongoing efforts in \textsl{using} the concept of visual attention in generative models \cite{Tang2013LearningGM, NIPS2018_7627, pmlr-v97-zhang19d}, but the focus of these methods is to use attention as an auxiliary information source for the particular task of interest, and not visually explain the generative model itself. 

In this work, we take a step towards bridging this crucial gap, developing new techniques to visually explain variational autoencoders (VAE) \cite{Kingma2013AutoEncodingVB}. Note that while we use VAEs as an instantiation of generative models in our work, some of the ideas we discuss are not limited to VAEs and can certainly be extended to GANs \cite{goodfellow2014generative} as well. Our intuition is that the latent space of a trained VAE encapsulates key properties of the VAE and that generating explanations conditioned on the latent space will help explain the reasoning for any downstream model predictions. Given a trained VAE, we present new ways to generate visual attention maps from the latent space by means of gradient-based attention. Specifically, given the learned Gaussian distribution, we use the reparameterization trick \cite{Kingma2013AutoEncodingVB} to sample a latent code. We then backpropagate the activations in each dimension of the latent code to a convolutional feature layer in the model and aggregate all the resulting gradients to generate the attention maps. While these visual attention maps serve as means to explain the VAE, we can do much more than just that. A classical application of a VAE is in anomaly localization, where the intuition is that any input data that is not from the standard Gaussian distribution used to train the VAE should be anomalous in the inferred latent space. Given this inference, we can now generate attention maps helping visually explain \textsl{why} this particular input is anomalous. We then also go a step further, presenting ways in which to use these explanations as cues to precisely localize where the anomaly is in the image. We conduct extensive experiments on the recently proposed MVTec anomaly detection dataset and present state-of-the-art anomaly localization results with just the standard VAE without any bells and whistles. 

Latent space disentanglement is another important area of study with VAEs and has seen much recent progress \cite{Higgins2017betaVAELB, Kim2018DisentanglingBF, Zheng_2019_CVPR}. With our visual attention explanations conditioned on the learned latent space, our intuition that using these attention maps as part of trainable constraints will lead to improved latent space disentanglement. To this end, we present a new learning objective we call attention disentanglement loss and show how one can train existing VAE models with this new loss. We demonstrate its impact in learning a disentangled embedding by means of experiments on the Dsprites dataset \cite{dsprites17}. 

To summarize, our key contributions are:

\begin{itemize}
    \item We take a step towards solving the relatively unexplored problem of visually explaining generative models, presenting new methods to generate visual attention maps conditioned on the latent space of a variational autoencoder. Furthermore, we show how our visual attention maps can be put to multipurpose use.
    \item We present new ways to localize anomalies in images by using our attention maps as cues, demonstrating state-of-the-art localization performance on the MVTec-AD dataset \cite{bergmann2019mvtec}.
    \item We present a new learning objective called the attention disentanglement loss, showing how one incorporate it into standard VAE models, and demonstrate improved disentanglement performance on the Dsprites dataset \cite{dsprites17}. 
\end{itemize}

\section{Related Work}\label{sec:work}

\textbf{CNN Visual Explanations.}
Much recent effort has been expended in explaining CNNs as they have come to dominate performance on most vision tasks. Some widely adopted methods that attempt to visualize intermediate CNN feature layers included the work of Zeiler and Fergus \cite{zeiler2014visualizing} and Mahendran and Vedaldi \cite{mahendran2015understanding}, where methods to understand the activity within the layers of convolutional nets were presented. Some more recent extensions of this line of work include visual-attention-based approaches \cite{CAM_CVPR16, ABN_CVPR19, GradCAM_ICCV17, gradCAMpp_WACV18}, most of which can be categorized into either gradient-based methods or response-based methods. Gradient-based method such as GradCAM \cite{GradCAM_ICCV17} compute and visualize gradients backpropagated from the decision unit to a feature convolutional layer. On the other hand, response-based approaches \cite{Zhang2016TopDownNA, CAM_CVPR16, ABN_CVPR19} typically add additional trainable units to the original CNN architecture to compute the attention maps. In both cases, the goal is to localize attentive and informative image regions that contribute the most to the model prediction. However, these methods and their extensions \cite{ABN_CVPR19, li2019guided, Wang_2019_ICCV}, while able to explain classification/categorization models, cannot be trivially extended to explaining deep generative models such as VAEs. In this work, we present methods, using the philosophy of gradient-based network attention, to compute and visualize attention maps directly from the learned latent embedding of the VAE. Furthermore, we make the resulting attention maps end-to-end trainable and show how such a change can result in improved latent space disentanglement. 

\textbf{Anomaly Detection.}
Unsupervised learning for anomaly detection \cite{akcay2018ganomaly} still remains challenging. Most recent work in anomaly detection is based on either classification-based \cite{ruff2018deep,chalapathy2018anomaly} or reconstruction-based approaches. Classification-based approaches aim to progressively learn representative one-class decision boundaries like hyperplanes \cite{chalapathy2018anomaly} or hyperspheres \cite{ruff2018deep} around the normal-class input distribution to tell outliers/anomalies apart. However, it was also shown \cite{chalapathy2019deep} that these methods have difficulty dealing with high-dimensional data. Reconstruction-based models, on the other hand, assume input data that are anomalous cannot be reconstructed well by a model that is trained only with normal input data. This principle has been used by several methods based on the traditional PCA \cite{kim2009observe}, sparse representation \cite{zhao2017spatio}, and more recently deep autoencoders \cite{zong2018deep,zhou2017anomaly}. In this work, we take a different approach to tackling this problem. We use the attention maps generated by our proposed VAE visual explanation generation method as cues to localize anomalies. Our intuition is that representations of anomalous data should be reflected in latent embedding as being anomalous, and that generating input visual explanations from such an embedding gives us the information we need to localize the particular anomaly.

\textbf{VAE Disentanglement.} 
Much effort has been expended in understanding latent space disentanglement for generative models. Early work of Schmidhuber \etal \cite{Schmidhuber:1992:LFC:159770.159784} proposed a principle to disentangle latent representations by minimizing the predictability of one latent dimension given other dimensions. Desjardins et. al \cite{Desjardins2012DisentanglingFO} generalized an approach based on restricted Boltzmann machines to factor the latent variables. Chen et. al extended GAN \cite{goodfellow2014generative} framework to design the InfoGAN \cite{NIPS2016_6399} to maximise the mutual information between a subset of latent variables and the observation. Some of the more recent unsupervised methods for disentanglement include $\beta$-VAE \cite{Higgins2017betaVAELB} which attempted to explore independent latent factors of variation in observed data. While still a popular unsupervised framework, $\beta$-VAE sacrificed reconstruction quality for obtaining better disentanglement. Chen et. al \cite{NIPS2018_7527} extended $\beta$-VAE to $\beta$-TCVAE by introducing a total correlation-based objective, whereas Mathieu \etal \cite{Mathieu2018DisentanglingD} explored decomposition of the latent representation into two factors for disentanglement, and Kim \etal \cite{Kim2018DisentanglingBF} proposed FactorVAE that encouraged the distribution of representations to be factorial and independent across the dimensions. While these methods focus on factorizing the latent representations provided by each individual latent neuron, we take a different approach. We enforce learning a disentangled space by formulating disentanglement constraints based on our proposed visual explanations, \ie, visual attention maps. To this end, we propose a new attention disentanglement learning objective that we quantitatively show provides superior performance when compared to existing work.

\section{Approach}
\label{sec:approach}
In this section, we present our method to generate explanations for a VAE by means of gradient-based attention. We first begin with a brief review of VAEs in Sections~\ref{sec:OneclassVAE} followed by our proposed method to generate VAE attention. We discuss our framework for localizing anomalies in images with these attention maps and conduct extensive experiments on the MVTec-AD anomaly detection dataset \cite{bergmann2019mvtec}, establishing state-of-the-art anomaly localization performance. Next, we show how our generated attention visualizations can assist in learning a disentangled latent space by optimizing our new attention disentanglement loss. Here, we conduct experiments on the Dsprites \cite{dsprites17} dataset and quantitatively demonstrate improved disentanglement performance when compared to existing approaches.

\subsection{One-Class Variational Autoencoder}
\label{sec:OneclassVAE}
A vanilla VAE is essentially an autoencoder that is trained with the standard autoencoder reconstruction objective between the input and decoded/reconstructed data, as well as a variational objective term attempts to learn a standard normal latent space distribution. The variational objective is typically implemented with Kullback-Leibler distribution metric computed between the latent space distribution and the standard Gaussian. Given input data $\mathbf{x}$, the conditional distribution $q(\mathbf{z}|\mathbf{x})$ of the encoder, the standard Gaussian distribution $p(\mathbf{z})$, and the reconstructed data $\hat{\mathbf{x}}$, the vanilla VAE optimizes:
\begin{equation}
    L=L_r(\mathbf{x},\hat{\mathbf{x}}) + L_{\text{KL}}(q(\mathbf{z}|\mathbf{x}),p(\mathbf{z}) )
    \label{eq:vaeloss}
\end{equation}
where $L_{\text{KL}}$ is the Kullback-Leibler divergence term and $L_r$ is the reconstruction term, which is typically a mean-squared error between $\mathbf{x}$ and $\hat{\mathbf{x}}$.


\subsection{Generating VAE Attention}
\begin{figure}[t]
\centering
\includegraphics[width=1\linewidth]{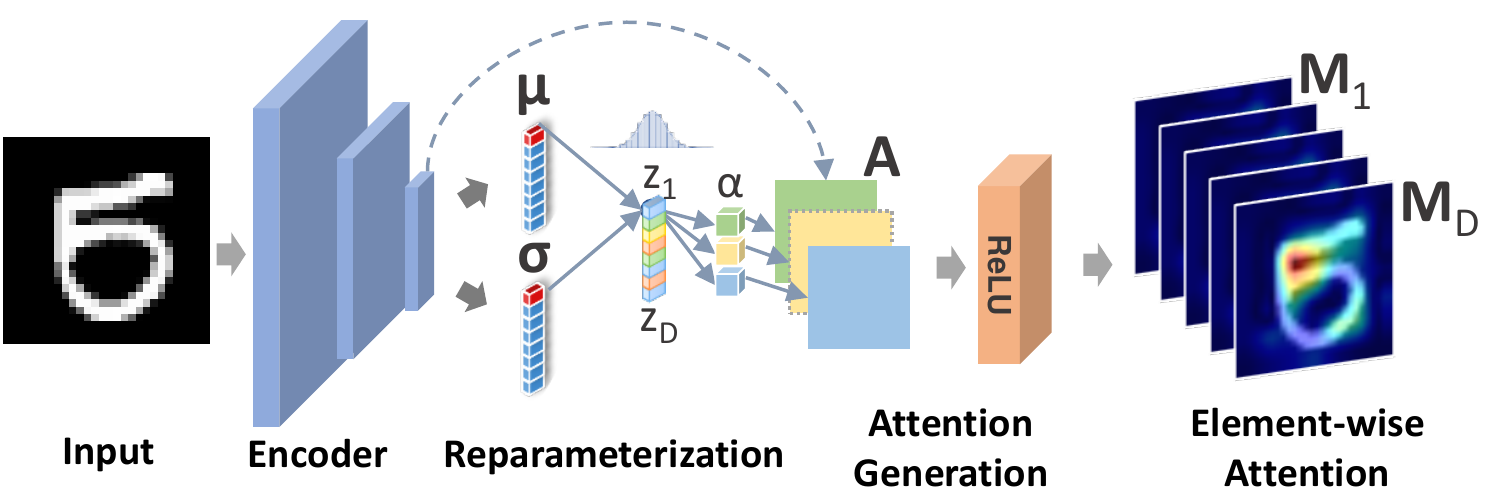}
\caption{Element-wise attention generation with a VAE.}
\label{fig:attention_gen}
\end{figure}

We propose a new technique to generate VAE visual attention by means of gradient-based attention computation. Our proposed approach is substantially different from existing work \cite{GradCAM_ICCV17, CAM_CVPR16, zheng2019re} that computes attention maps by backpropagating the score from a classification model. On the other hand, we are not restricted by such requirements and develop an attention mechanism directly using the learned latent space, thereby not needing an additional classification module. As illustrated in Figure~\ref{fig:attention_gen} and discussed below, we compute a score from the latent space, which is then used to calculate gradients and obtain the attention map.

Specifically, given the posterior distribution $q(\mathbf{z}|\mathbf{x})$ inferred by the trained VAE for a data sample $\mathbf{x}$, we use the reparameterization trick to obtain a latent vector $\textbf{z}$. For each  element $z_{i}$, we backpropagate gradients to the last convolutional feature maps $\mathbf{A} \in \mathbb{R}^{n\times h\times w}$, giving the attention map $\mathbf{M}^{i}$ corresponding to $z_{i}$. Specifically, $\mathbf{M}^{i}$ is computed as the linear combination: 
\begin{equation}
\mathbf{M}^{i}=\text{ReLU}(\sum_{k=1}^{n}\alpha_{k}\mathbf{A}_{k})
\label{eq:att_zi}
\end{equation} 
where the scalar $\alpha_{k}=\text{GAP}(\frac{\partial z_{i}}{\partial \mathbf{A}_{k}})$ and $\mathbf{A}_{k}$ is the $k^{th}$ feature channel ($k=1,\ldots,n$) of the feature maps $\mathbf{A}$. Note $\frac{\partial z_{i}}{\partial \mathbf{A}_{k}}$ is a matrix and so we use the global average pooling (GAP) operation to get the scalar $\alpha_{k}$. Specifically, this is: 
\begin{equation}
\alpha_{k}=\frac{1}{T}\sum_{p=1}^{h}\sum_{q=1}^{w}(\frac{\partial z_{i}}{\partial A^{pq}_{k}})
\end{equation}
where $T=h\times w$ and $A^{pq}_{k}$ is the pixel value at location $(p,q)$ of the $h\times w$ matrix $\mathbf{A}_{k}$. We now repeat this for all elements $z_{1}, z_{2}, \ldots, z_{D}$ of the $D-$dimensional latent space, giving $\mathbf{M}^{1}, \ldots, \mathbf{M}^{D}$ (see Figure~\ref{fig:attention_gen}). An example of what each $\mathbf{M}^{i}$ represents is shown in Figure~\ref{fig:teaser}, where we see consistent high-response regions for each latent dimension across multiple data samples. While the above procedure gives one attention map per latent dimension, one can obtain a single overall attention map using any matrix aggregation scheme, \eg, the mean, in which case the overall attention map is $\mathbf{M} = \frac{1}{D} \sum_i^D \mathbf{M}^{i}$.

\subsection{Generating Anomaly Attention Explanations}
\begin{figure*}[t]
\centering
\includegraphics[width=1\linewidth]{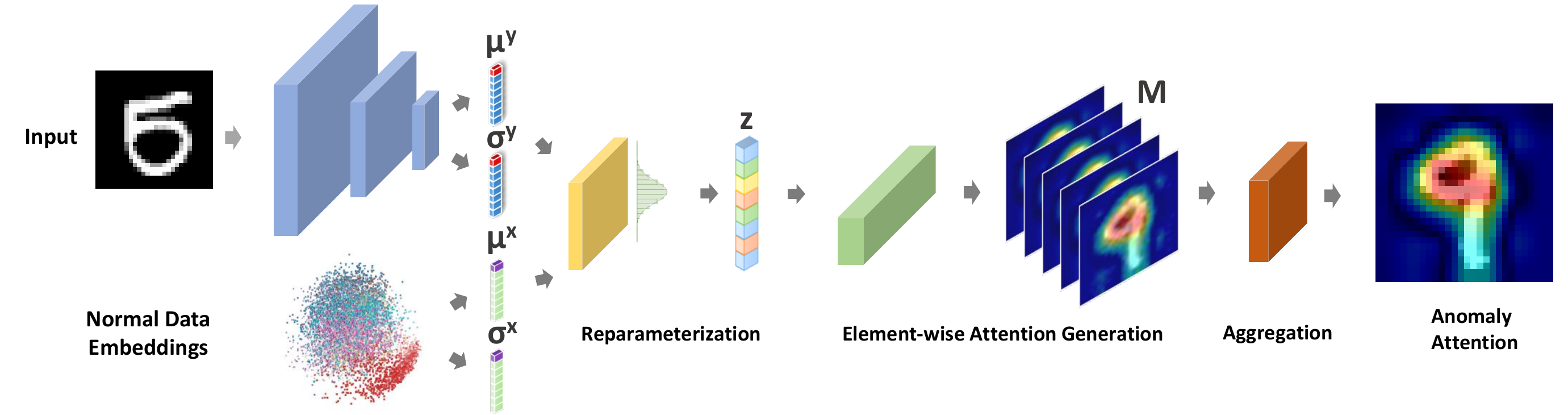}
\caption{Attention generation with a one-class VAE.}
\label{fig:oneclassAttentionAnomaly}
\end{figure*}

We now discuss how our gradient-based attention generation mechanism can be used to localize anomaly regions given a trained one-class VAE. Inference with such a one-class VAE with data it was trained for, \ie, normal data (digit ``1" for instance), should ideally result in the learned latent space representing the standard normal distribution. Consequently, given a testing sample from a different class (abnormal data, digit ``5" for instance), the latent representation inferred by the learned encoder should have a large difference when compared to the learned normal distribution. This intuition can be captured in many ways. A straightforward approach (which we use to show results next) is to take the inferred mean vector and generate the resulting attention map. Specifically, we compute the sum of all elements in the mean vector, giving a score $s$, which we backpropagate to compute the anomaly attention $\mathbf{M}$ (as in Equation~\ref{eq:att_zi}). An alternative approach can be using the normal difference distribution. Given all normal images used to train the VAE, we can infer the overall $\mu^x$ and $\sigma^x$ representing the distribution of embeddings of all the normal images $\mathbf{x}\in \mathbf{X}$. Now, given  the $\mathbf{\mu}^{y}_i$ and $\mathbf{\sigma}^{y}_i$ for each latent variable $z_i$ inferred for an abnormal sample $\mathbf{y}$, we can define the normal difference distribution as: 
\begin{equation}
    P_{q(z_i|x)-q(z_i|y)}(u) = \frac{e^{-[u-(\mu^{x}_i-\mu^{y}_i)]^2 /[2((\sigma^{x}_i)^2+(\sigma^{y}_i)^2)]}}{\sqrt{2\pi((\sigma^{x}_i)^2+(\sigma^{y}_i)^2)}}
    \label{ndd}
\end{equation}
for each latent variable $z_i$. Given a latent code $\mathbf{z}$ sampled from $P_{q(z_i|X)-q(z_i|Y)}$, one can follow the procedure described above to compute the anomaly attention map $\mathbf{M}$. This is visually summarized in Figure~\ref{fig:oneclassAttentionAnomaly}.

\begin{figure}[h]
\includegraphics[width=\linewidth]{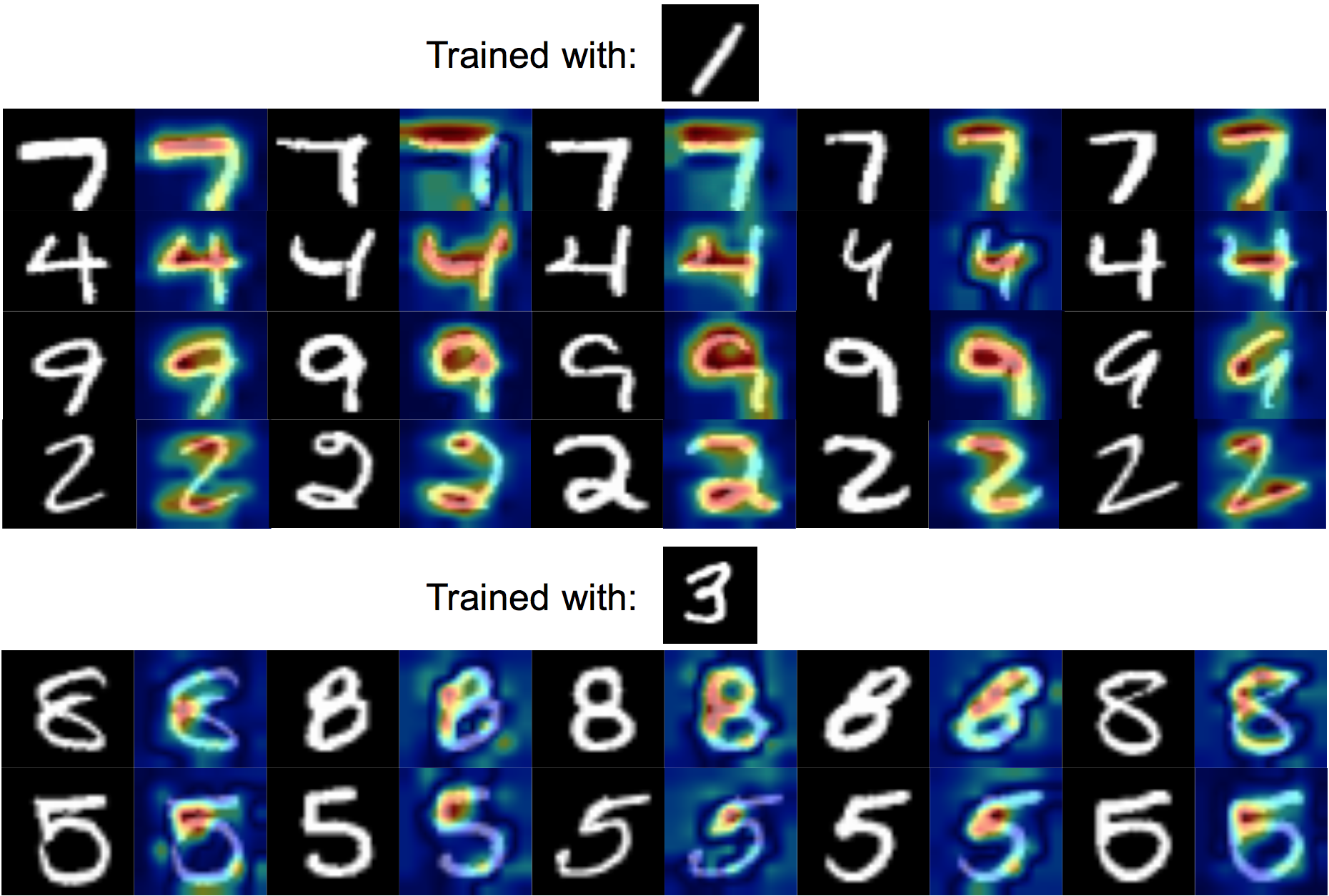}
\caption{Anomaly localization results from the MNIST dataset.}
\label{fig:mnist}
\end{figure}

\begin{figure}[h]
\includegraphics[width=\linewidth]{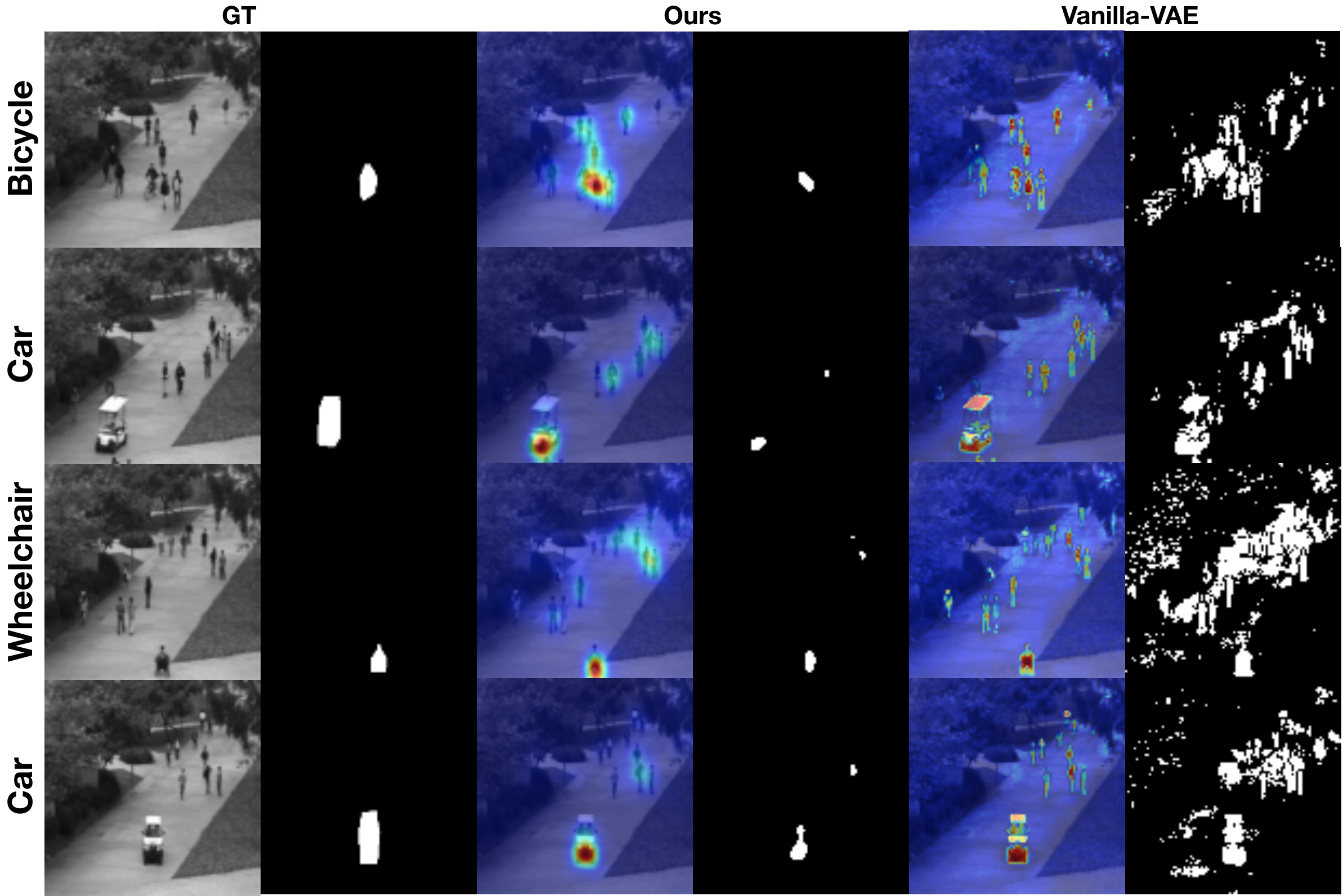}
\caption{Qualitative results from UCSD Ped1 dataset. L-R: Original test image, ground-truth masks, our anomaly attention localization maps, and difference between input and the VAE's reconstruction
. The anomalies in these samples are moving cars, bicycle, and wheelchair.} 
\label{fig:ucsd}
\end{figure}

\subsubsection{Results}
In this section, we evaluate our proposed method to generate visual explanations as well as perform anomaly localization with VAEs.

\noindent {\bf Metrics:} We adopt the commonly used the area under the receiver operating characteristic curve (ROC AUC) for all quantitative performance evaluation. We define true positive rate (TPR) as the percentage of pixels that are correctly classified as anomalous across the whole testing class, whereas the false positive rate (FPR) the percentage of pixels that are wrongly classified as anomalous. In addition, we also compute the best intersection-over-union (IOU) score by searching for the best threshold based on our ROC curve. Note that we first begin with qualitative (visual) evaluation on the MNIST and UCSD datasets, and then proceed to a more thorough quantitative evaluation on the MVTec-AD dataset. 

\noindent {\bf MNIST.} We start by qualitatively evaluating our visual attention maps on the MNIST dataset \cite{deng2012mnist}. Using training images from one single digit class, we train our one-class VAE model, which will be used to test on all the digit numbers' testing images. We reshape all the training and testing images to resolution of $28\times28$ pixels. 

In Figure~\ref{fig:mnist} (top), we show results with a model trained on  the digit ``1" (normal class) and test on all other digits (each of which becomes an abnormal class). For each test image, we infer the latent vector using our trained encoder and generate the attention map. As can be observed in the results, the attention maps computed with the proposed method is intuitively satisfying. For instance, let us consider the attention maps generated with digit ``7" as the test image. Our intuition tells us that a key difference between the ``1" and the ``7" is the top-horizontal bar in ``7", and our generated attention map indeed highlights this region. Similarly, the differences between an image of the digit ``2" and the ``1" are the horizontal base and the top-round regions in the ``2". From the generated attention maps for ``2", we notice that we are indeed able to capture these differences, highlighting the top and bottom regions in the images for ``2". We also show testing results with other digits (\eg, ``4",``9") as well as with a model trained on digit ``3" and tested on the other digits in the same figure. We note similar observations can be made from these results as well, suggesting that our proposed attention generation mechanism is indeed able to highlight anomalous regions, thereby capturing the features in the underlying latent space that cause a certain data sample to be abnormal.

\noindent {\bf UCSD Ped1 Dataset:} We next test our proposed method on the UCSD Ped 1\cite{li2013anomaly} pedestrian video dataset, where the videos were captured with a stationary camera to monitor a pedestrian walkway. This dataset includes 34 training sequences and 36 testing sequences, with about 5500 ``normal" frames and 3400 ``abnormal" frames. We resize the data to $100\times 100$ pixels for training and testing. 

We first qualitatively evaluate the performance of our proposed attention generation method in localizing anomalies. As we can see from Figure~\ref{fig:ucsd} (where the corresponding anomaly of interest is annotated on the left, \eg, \textsl{bicycle}, \textsl{Car} \etc), our anomaly localization technique with attention maps performs substantially better than simply computing the difference between the input and its reconstruction (this result is annotated as \textsl{Vanilla-VAE} in the figure). We note more precise localization of the high-response regions in our generated attention maps, and these high-response regions indeed correspond to anomalies in these images.

We next conduct a simple ablation study using the pixel-level segmentation AUROC score against the baseline method of difference between input data and the reconstruction. We test our proposed attention generation mechanism with varying levels of spatial resolution by backpropagating to each of the encoder's convolutional layers: $50\times50$, $25\times25$, and $12\times12$. The results are shown in Table~\ref{tab:ucsd} where we see our proposed mechanism gives better performance than the baseline technique. 

\begin{table}[h]
    \centering
    \resizebox{\columnwidth}{!}{
    \begin{tabular}{c|c|c|c|c}
    \hline
    & Vanilla-VAE & Ours(Conv1) & Ours(Conv2) & Ours(Conv3) \\
    \hline
    AUROC  & 0.86 & 0.89 & \bf{0.92} & 0.91 \\
    \hline
    \end{tabular}
    }
    \caption{Results on UCSD Ped1 using pixel-level segmentation AUROC score. We compare results obtained using our anomaly attention generated with different target network layers to reconstruction-based anomaly localization using Vanilla-VAE. }
    \label{tab:ucsd}
\end{table}

\noindent {\bf MVTec-AD Dataset:} We consider the recently released comprehensive anomaly detection dataset: MVTec Anomaly Detection (MVTec AD) \cite{bergmann2019mvtec} that provides multi-object, multi-defect natural images and pixel-level ground truth. This dataset contains 5354 high-resolution color images of different objects/textures, with both normal and defect (abnormal) images provided in the testing set. We resize all images to $256\times256$ pixels for training and testing. We conduct extensive qualitative and quantitative experiments and summarize results below. 

We train a VAE with ResNet18 \cite{he2016deep} as our feature encoder and a 32-dimensional latent space. We further use random mirroring and random rotation, as done in the original work \cite{bergmann2019mvtec}, to generate an augmented training set. Given a test image, we infer its latent representation $\mathbf{z}$ to generate the anomaly attention map. Given our anomaly attention maps, we generate binary anomaly localization maps using a variety of thresholds on the pixel response values, which is encapsulated in the ROC curve. We then compute and report the area under the ROC curve (ROC AUC) and generate the best IOU number for our method based on FPR and TPR from the ROC curve. 

The results are shown in Table~\ref{tab:mvtec}, where we compare our performance with the techniques evaluated in the benchmark paper of Bergmann \etal \cite{bergmann2019mvtec} (note that the baselines here are the same methods as in \cite{bergmann2019mvtec}). From the results, we note that with our anomaly localization approach using the proposed VAE attention, we obtain better results on most of the object categories than the competing methods. It is worth noting here that some of these methods are specifically designed for the anomaly localization task, whereas we train a standard VAE and generate our VAE attention maps for localization. Despite this simplicity, our method achieves competitive performance, demonstrating the potential of such an attention generation technique to be useful for tasks other than just model explanation.

\begin{figure}[h]
\includegraphics[width=\linewidth]{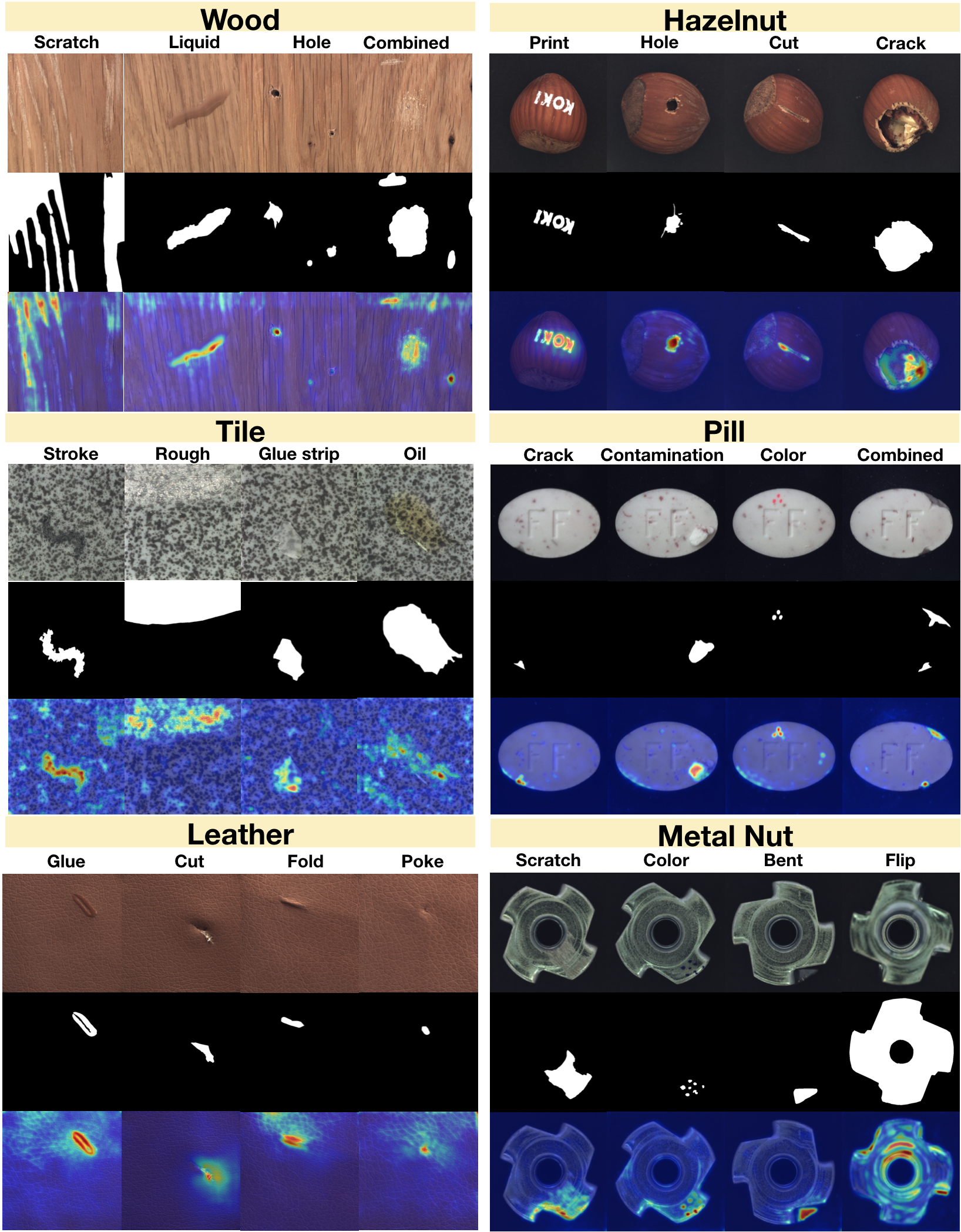}
\caption{Qualitative results from MVTec-AD. Here, we provide results from: Wood, Tile, Leather, Hazelnut, Pill, and Metal Nut. For each category, we show four different type of defects. As can be seen from the figure, our anomaly attention maps are able to accurately localize anomalies. }
\label{fig:mvtec}
\end{figure}

We also show some qualitative results in Figure~\ref{fig:mvtec}. We show results from six categories - three textures and three objects. For each category, we also show four types of defects provided by the dataset. We show, from the top row to the bottom, the original images, ground truth segmentation masks, and our anomaly attention maps. One can note that our attention maps are able to accurately localize anomalous regions across these various defect categories.


\begin{table}[h]
\centering
\resizebox{\columnwidth}{!}{ 
\begin{tabular}{cccccc|c}
\hline
\multicolumn{2}{c}{Category} & \begin{tabular}[c]{@{}c@{}}AE\\(SSIM)\end{tabular} & \begin{tabular}[c]{@{}c@{}}AE\\(L2)\end{tabular}  &  \begin{tabular}[c]{@{}c@{}}Ano\\GAN\end{tabular} &  \begin{tabular}[c]{@{}c@{}c@{}}CNN\\ Feature\\Dictionary\end{tabular}  &  ours \\ \hline \hline
\parbox[t]{2mm}{\centering \multirow{10}{*}{\rotatebox[origin=c]{90}{Texture}}} & 
\parbox[t]{10mm}{\centering \multirow{2}{*}{\rotatebox[origin=c]{0}{Carpet}}} &\bf{0.87} &0.59 &0.54 &0.72 & 0.78\\
 & & \bf{0.69} & 0.38 & 0.34 & 0.20  & 0.1\\ \cline{2-7}
 &\parbox[c]{10mm}{\multirow{2}{*}{Grid}} &\bf{0.94} & 0.90 & 0.58 &0.59  & 0.73\\ 
 & & \bf{0.88} & 0.83 & 0.04 & 0.02 & 0.02 \\ \cline{2-7}
  & \parbox[c]{10mm}{\multirow{2}{*}{Leather}} & 0.78 & 0.75 &0.64 & 0.87 & \bf {0.95} \\ 
  & & 0.71 & 0.67& 0.34 & \bf{0.74} & 0.24 \\ \cline{2-7}
  & \parbox[c]{10mm}{\multirow{2}{*}{Tile}} & 0.59 &0.51 & 0.50& \bf{0.93} &0.80 \\
  & & 0.04 & \bf{0.23} & 0.08 & 0.14  & \bf{0.23} \\\cline{2-7}
  &\parbox[c]{10mm}{\multirow{2}{*}{Wood}} &0.73 &0.73 & 0.62& \bf{0.91} & 0.77\\ 
  & & 0.36 & 0.29 & 0.14 & \bf{0.47}  & 0.14 \\\hline
\parbox[c]{2mm}{\multirow{20}{*}{\rotatebox[origin=c]{90}{Objects}}}& 
\parbox[c]{10mm}{\multirow{2}{*}{Bottle}} & \bf{0.93}& 0.86 & 0.86 &0.78&0.87\\
& & 0.15 &0.22& 0.05& 0.07&\bf{0.27}\\ \cline{2-7}
& \parbox[c]{10mm}{\multirow{2}{*}{Cable}} &0.82 & 0.86& 0.78 & 0.79  & \bf{0.90}\\
& & 0.01& 0.05& 0.01& 0.13  & \bf{0.18} \\ \cline{2-7}
& \parbox[c]{10mm}{\multirow{2}{*}{Capsule}} &\bf{0.94} &0.88& 0.84& 0.84& 0.74\\
& & 0.09 & \bf{0.11} & 0.04 & 0.00 & \bf{0.11} \\ \cline{2-7}
& \parbox[c]{10mm}{\multirow{2}{*}{Hazelnut}} &0.97 &0.95& 0.87&0.72& \bf{0.98}\\
& & 0.00 & 0.41 & 0.02& 0.00 &  \bf{0.44} \\\cline{2-7}
& \parbox[c]{10mm}{\multirow{2}{*}{Metal Nut}} & 0.89 & 0.86 & 0.76&0.82  & \bf{0.94}\\
& & 0.01 & 0.26 & 0.00 & 0.13  & \bf{0.49}\\\cline{2-7}
& \parbox[c]{10mm}{\multirow{2}{*}{Pill}} &\bf{0.91}& 0.85& 0.87&  0.68& 0.83\\
& & 0.07 & \bf{0.25}& 0.17 & 0.00 & 0.18\\\cline{2-7}
& \parbox[c]{10mm}{\multirow{2}{*}{Screw}} &0.96 & 0.96& 0.80 &0.87  & \bf{0.97}\\
& & 0.03 & \bf{0.34} & 0.01 & 0.00 & 0.17\\\cline{2-7}
& \parbox[c]{10mm}{\multirow{2}{*}{Toothbrush}}&0.92 &0.93& 0.90& 0.77& \bf{0.94}\\
& & 0.08 & \bf{0.51} & 0.07 & 0.00& 0.14\\\cline{2-7}
& \parbox[c]{10mm}{\multirow{2}{*}{Transistor}}& 0.90 &0.86& 0.80& 0.66 & \bf{0.93}\\
& & 0.01& 0.22 & 0.08 & 0.03& \bf{0.30} \\\cline{2-7}
& \parbox[c]{10mm}{\multirow{2}{*}{Zipper}} &\bf{0.88} &0.77 &0.78& 0.76 & 0.78\\ 
& & 0.10 & \bf{0.13} & 0.01 & 0.00  & 0.06 \\\hline
\end{tabular}
}
\caption{Quantitative results for pixel level segmentation on 15 categories from MVTec-AD dataset. For each category, we report the area under ROC AUC curve on the top row, and best IOU on the bottom row. We adopt comparison scores from \cite{bergmann2019mvtec}. }
\label{tab:mvtec}
\end{table}

\subsection{Attention Disentanglement} \label{subsec:ad-factorvae}
In the previous section, we discussed how one can generate visual explanations, by means of gradient-based attention, as well as anomaly attention maps for VAEs. We also discussed and experimentally evaluated using these anomaly attention maps for anomaly localization on a variety of datasets. We next discuss another application of our proposed VAE attention: VAE latent space disentanglement. Existing approaches for learning disentangled representations of deep generative models focus on formulating factorised, independent latent distributions so as to learn interpretable data representations. Some examples include $\beta$-VAE \cite{Higgins2017betaVAELB}, InfoVAE \cite{Zhao2017InfoVAEIM}, and FactorVAE \cite{Kim2018DisentanglingBF}, among others, all of which attempt to model the latent prior with factorial probability distribution. In this work, we present an alternative technique, based on our proposed VAE attention, called the attention disentanglement loss. We show how it can be integrated with existing baselines, \eg, FactorVAE, and demonstrate the resulting impact by means of qualitative attention maps and quantitatively performance characterization with standard disentanglement metrics.

\begin{figure}[h]
\centering
\includegraphics[width=1\linewidth]{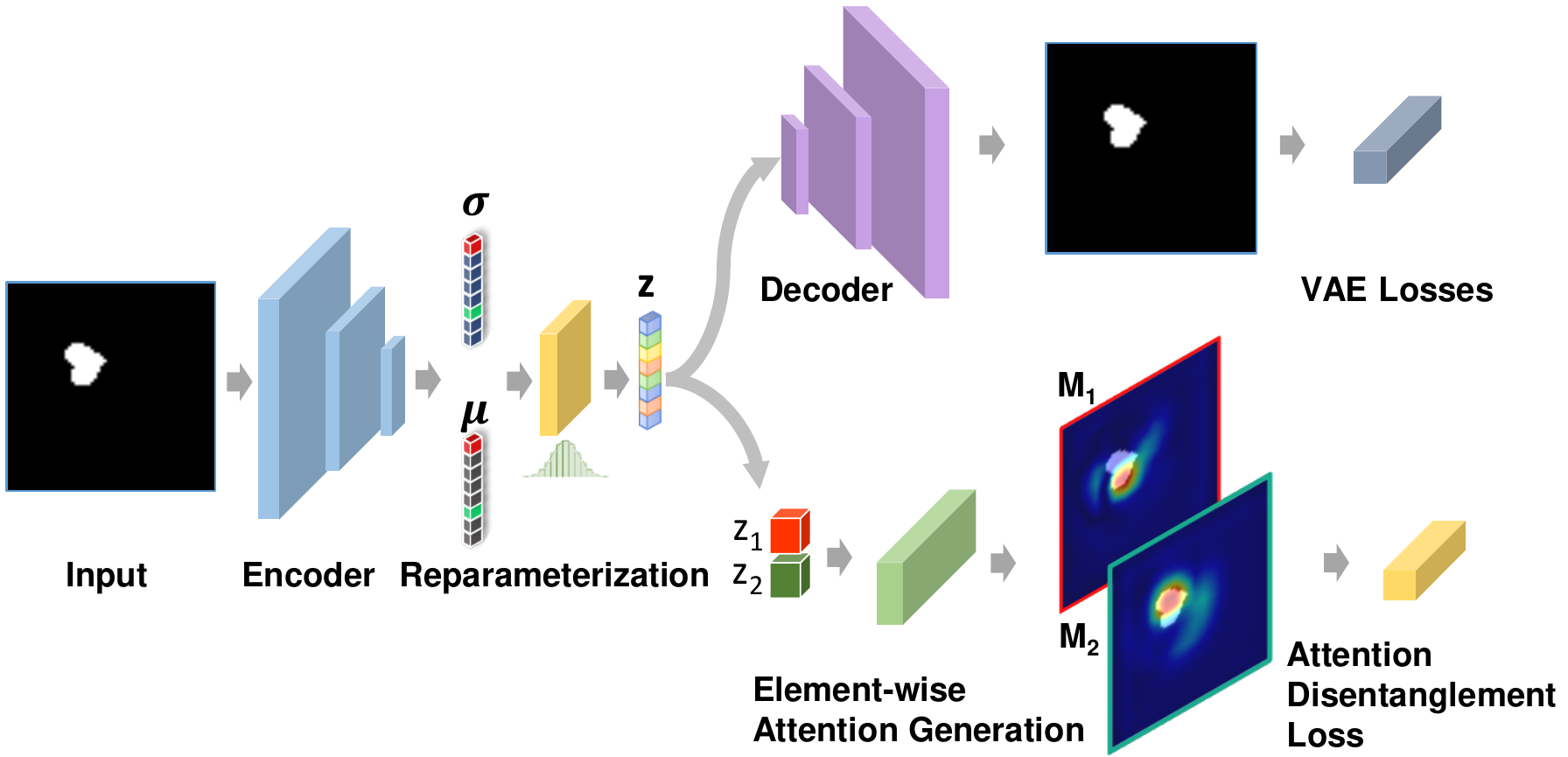}
\caption{Training a variational autoencoder with the proposed attention disentanglement loss.}
\label{fig1}
\end{figure}
    
\subsubsection{Training with Attention Disentanglement}
As we showed earlier, our proposed VAE attention, by means of gradient-based attention, generates attention maps that can explain the underlying latent space represented by the trained VAE. We showed how attention maps intuitively represent different regions of normal and abnormal images, directly corresponding to differences in the latent space (since we generate attention from the latent code).  Consequently, our intuition is that using these attention maps to further bootstrap the training process of the VAE model should help boost latent space disentanglement. To this end, our big-picture idea is to use these attention maps as trainable constraints to explicitly force the attention computed from the various dimensions in latent space to be as disentangled, or separable as possible. Our hypothesis is that if we are able to achieve this, we will be able to learn an improved disentangled latent space. To realize this objective, we propose a new loss called the \textsl{attention disentanglement} loss ($L_{\text{AD}}$) that can be easily integrated with existing VAE-type models (see Figure~\ref{fig1}). Note that while we use the FactorVAE \cite{Kim2018DisentanglingBF} for demonstration in this work, the proposed attention disentanglement loss is in no way limited to this model and can be used in conjunction with other models as well (\eg, $\beta$-VAE \cite{Higgins2017betaVAELB}). The proposed $L_{\text{AD}}$ takes two attention maps $\mathbf{A}^1$ and $\mathbf{A}^2$ (each computed from a certain dimension in the latent space following Equation~\ref{eq:att_zi}) as input, and attempts to separate the high-response pixel regions in them as much as possible. This can be mathematically expressed as: 

\begin{equation}
L_{AD} = 2\cdot \frac{\sum_{ij}min(A_{ij}^1, A_{ij}^2)}{\sum_{ij}A_{ij}^1+ A_{ij}^2}\label{dis:eq3}
\end{equation}
where $\cdot$ is the scalar product operation, and $A_{ij}^1$ and $A_{ij}^2$ are the $(i,j)^{th}$ pixel in the attention maps $\mathbf{A}^1$ and $\mathbf{A}^2$ respectively. The proposed $L_{AD}$ can be directly integrated with the standard FactorVAE training objective $L_{\text{FV}}$, giving us an overall learning objective that can be expressed as:

\begin{equation}
L = L_{\text{FV}}+\lambda L_{\text{AD}}
\label{dis:overall}
\end{equation}

We now train the FactorVAE with our proposed overall learning objective of Equation~\ref{dis:overall}, and evaluate the impact of $L_{\text{AD}}$ by comparisons with the baseline FactorVAE trained only with $L_{\text{FV}}$. For this purpose, we use the same evaluation metric discussed in FactorVAE \cite{Kim2018DisentanglingBF}.

\subsubsection{Results}

\noindent {\bf Data:} We use the Dsprites dataset \cite{dsprites17} which provides 737,280 binary $64\times64$ 2D shape images. \\

\begin{figure}
    \centering
    \includegraphics[width=\linewidth]{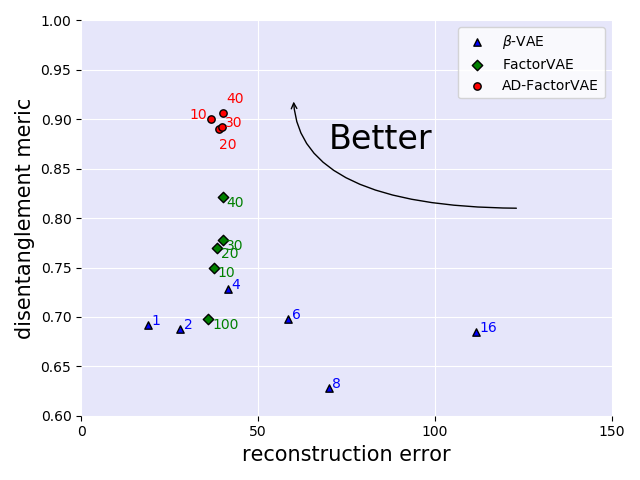}
    \caption{Reconstruction error plotted against disentanglement metric \cite{Kim2018DisentanglingBF}. The numbers at each point show $\beta$ and $\gamma$ values. We want a low reconstruction error and a high disentanglement metric.}
    \label{fig:dis_eval_metric}
\end{figure}

\noindent {\bf Quantitative Results:}
In Figure~\ref{fig:dis_eval_metric}, we compare the best disentanglement performance (plotted against the reconstruction error) of our proposed method (called AD-FactorVAE) with other competing approaches:  baseline FactorVAE \cite{Kim2018DisentanglingBF} (training with only $L_{\text{FV}}$) and $\beta$-VAE\cite{Higgins2017betaVAELB}. We note that training with our proposed $L_{\text{AD}}$ results in higher disentanglement scores under the same experimental setting, giving a best disentanglement score of around 0.90, whereas baseline FactorVAE ($\gamma=40$) gives around 0.82, both with a reconstruction error around 40. We also note our proposed method obtains a higher disentanglement score compared to $\beta$-VAE (0.73 with $\beta=4$ as the best result). These results demonstrate the potential of both our proposed VAE attention and $L_{\text{AD}}$ in improving the performance of existing methods in the disentanglement literature. These improvements are also reflected in the qualitative results we discuss next. 

\begin{figure}
    \centering
    \includegraphics[width=\linewidth]{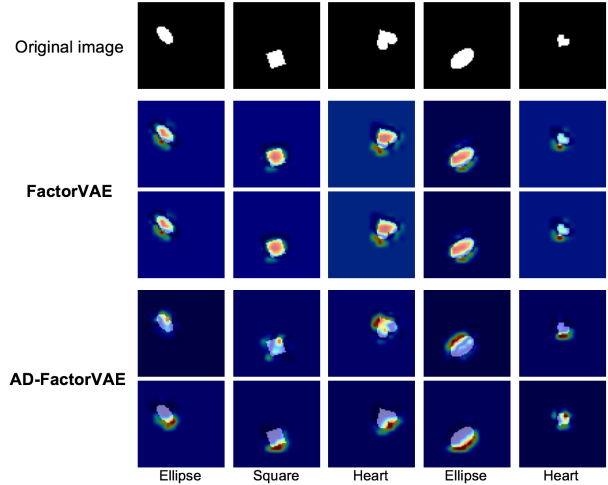}
    \caption{Attention separation on the Dsprites dataset. Top row: the original shape images. Middle two rows: attention maps from FactorVAE. Bottom two rows: attention maps from AD-FactorVAE.}
    \label{fig:dis_qua}
\end{figure}

\noindent {\bf Qualitative Results:}
Figure~\ref{fig:dis_qua} shows some attention maps generated using the baseline FactorVAE and our proposed AD-FactorVAE. The first row shows 5 input images, and the next 4 rows show results with the baseline FactorVAE and our proposed method. Row 2 shows attention maps generated with FactorVAE by backpropagating from the latent dimension with the highest response, whereas row 3 shows attention maps generated by backpropagating from the latent dimension with the next highest response. Rows 4 and 5 show the corresponding attention maps with the proposed AD-FactorVAE. Our intuition and expectation with AD-FactorVAE is that each dimension’s attention map will have high responses in different spatial regions of the input. From Figure~\ref{fig:dis_qua}, this is indeed the case, with high-response regions in different areas in the image (rows 4 and 5), whereas we see attention overlap in baseline FactorVAE (rows 2 and 3). 

\section{Summary and Future Work}
\label{sec:conclusion}
We presented new techniques to visually explain variational autoencoders, taking a first step towards explaining deep generative models by means of gradient-based network attention. We showed how one can use the learned latent representation to compute gradients and generate VAE attention maps, without relying on classification-kind of models. We demonstrating applicability of the resulting VAE attention on two tasks: anomaly localization and latent space disentanglement. In anomaly localization, we used the fact that an abnormal input will result in latent variables that do not conform to the standard Gaussian in gradient backpropagation and attention generation. These anomaly attention maps were then used as cues to generate pixel-level binary anomaly masks. In latent space disentanglement, we showed how we can use our VAE attention from each latent dimension to enforce new attention disentanglement learning constraints, resulting in improved attention separability as well as disentanglement performance. Since a VAE can infer a full posterior distribution, with our method, one can obtain a distribution of attention matrices (maps) with repeated sampling. While one way of visualizing this distribution is with the resulting sample mean, generating more generic visual explanations for the full matrix distribution is an interesting topic for future research.

\section*{Acknowledgements}
This material is based upon work supported in part by NSF grants 1911197, IIS–1814631, ECCS–1808381 and CMMI–1638234, and the U.S. Department of Homeland Security, Science and Technology Directorate, Office of University Programs, under Grant Award 2013-ST-061-ED0001. The views and conclusions contained in this document are those of the authors and should not be interpreted as necessarily representing the official policies, either expressed or implied, of the U.S. Department of Homeland Security.

{\small
\bibliographystyle{ieee_fullname}
\bibliography{egbib}
}

\end{document}